\documentclass[final]{cvpr}

\usepackage{times}
\usepackage{epsfig}
\usepackage{graphicx}
\usepackage{amsmath}
\usepackage{amssymb}
\usepackage{float}
\usepackage{booktabs}
\usepackage{boldline}
\usepackage{multirow}
\usepackage{color}
\usepackage{arydshln}

\usepackage[pagebackref=true,breaklinks=true,colorlinks,bookmarks=false]{hyperref}

\def\cvprPaperID{1129} 
\def\confYear{CVPR 2021}

\begin{document}

\title{The Spatially-Correlative Loss for Various Image Translation Tasks}

\author{Chuanxia Zheng \qquad Tat-Jen Cham\\
School of Computer Science and Engineering\\
Nanyang Technological University, Singapore \\
{\tt\small \{chuanxia001,astjcham\}@ntu.edu.sg}
\and
Jianfei Cai\\
Department of Data Science \& AI\\
Monash University, Australia\\
{\tt\small Jianfei.Cai@monash.edu}
}

	\maketitle


\begin{abstract}
	
We propose a novel spatially-correlative loss that is simple, efficient and yet effective for preserving scene structure consistency while supporting large appearance changes during unpaired image-to-image (I2I) translation. Previous methods attempt this by using pixel-level cycle-consistency or feature-level matching losses, but the domain-specific nature of these losses hinder translation across large domain gaps. To address this, we exploit the spatial patterns of self-similarity as a means of defining scene structure. Our spatially-correlative loss is geared towards only capturing spatial relationships within an image rather than domain appearance. We also introduce a new self-supervised learning method to explicitly learn spatially-correlative maps for each specific translation task. We show distinct improvement over baseline models in all three modes of unpaired I2I translation: single-modal, multi-modal, and even single-image translation. This new loss can easily be integrated into existing network architectures and thus allows wide applicability. The code is available at \href{https://github.com/lyndonzheng/F-LSeSim}{https://github.com/lyndonzheng/F-LSeSim}. 
   
\end{abstract}

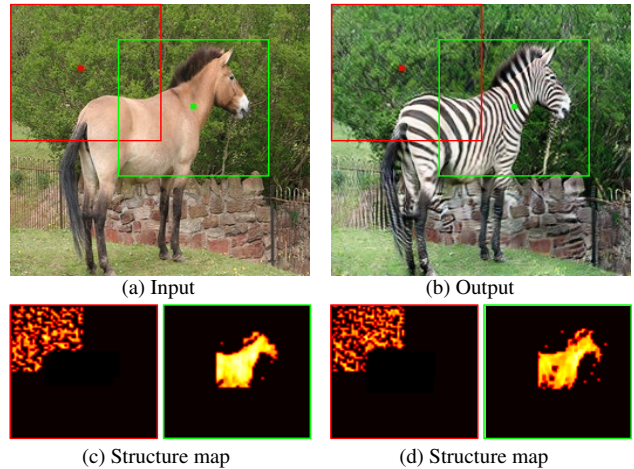
\begin{figure}[tb!]
	\centering
	\includegraphics[width=\linewidth,height=0.26\textheight]{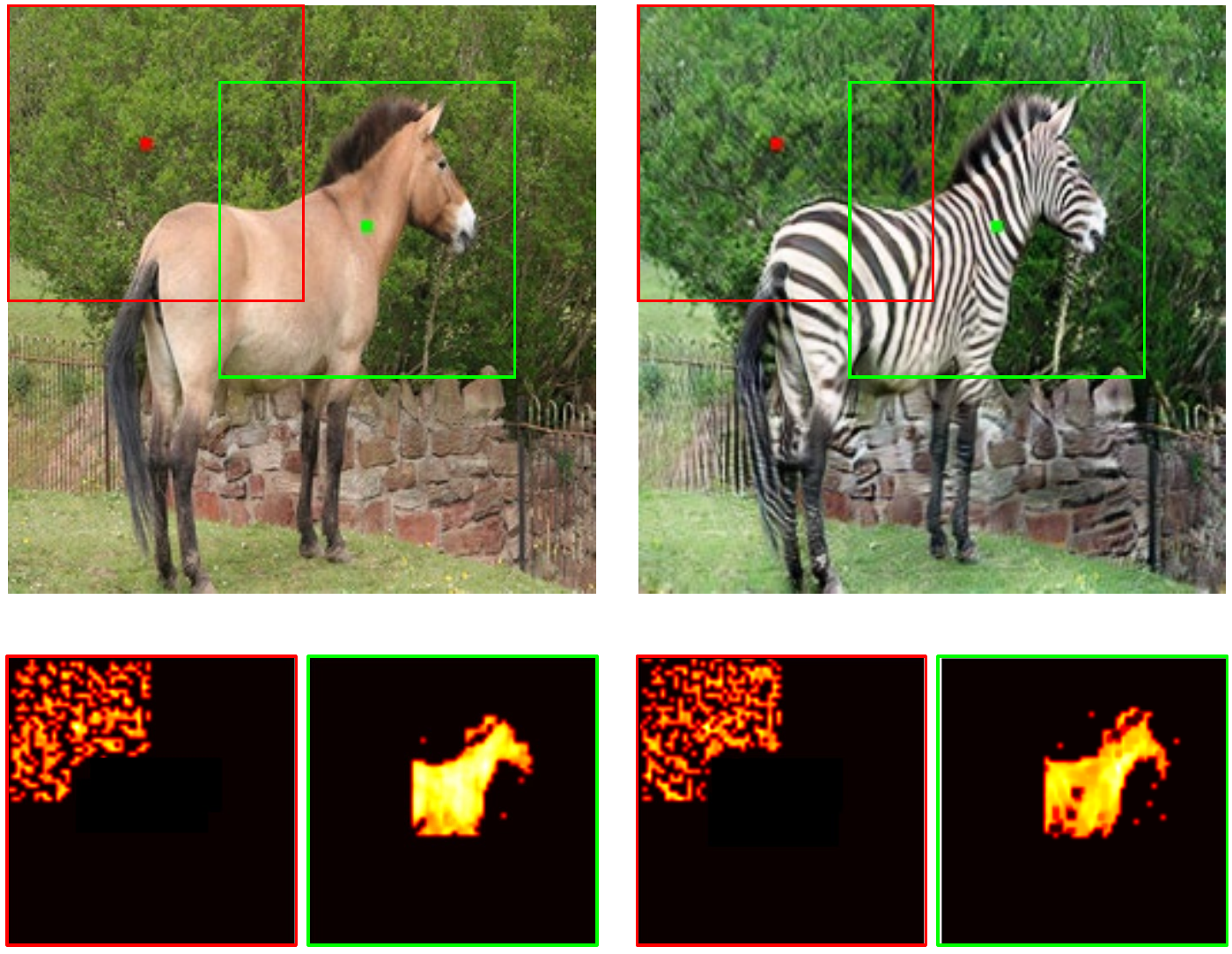}
	\begin{picture}(0,0)
	\put(-75,68){\footnotesize (a) Input}
	\put(40,68){\footnotesize (b) Output}
	\put(-90,4){\footnotesize (c) Structure map}
	\put(30,4){\footnotesize (d) Structure map}
	\end{picture}
	\caption{\textbf{Our learned spatially-correlative representation} encodes local scene structure based on self-similarities. Despite vast appearance differences between the \emph{horse} and \emph{zebra}, when the scene structures are identical (\ie same poses), the spatial patterns of self-similarities are as well.} 
	\label{fig:example}
\end{figure}

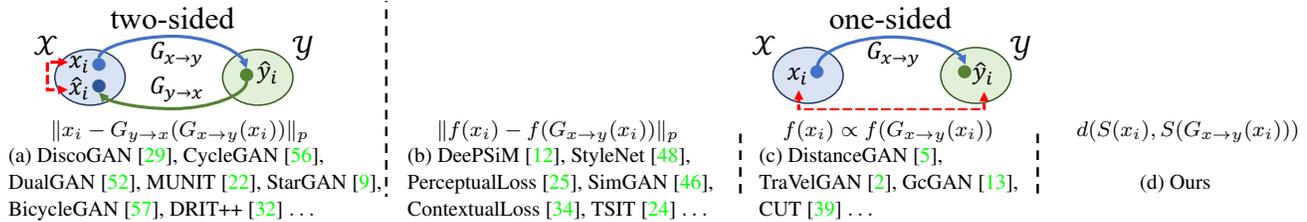
\begin{figure*}[tb!]
	\centering
	\setlength{\abovecaptionskip}{0.cm}
	\setlength{\belowcaptionskip}{-0.cm}
	\includegraphics[width=\linewidth,height=0.115\textheight]{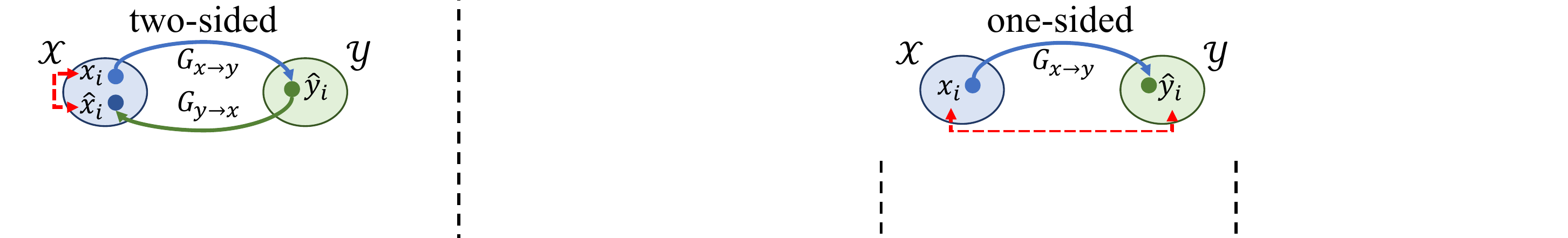}
	\begin{picture}(0,0)
		\put(-230,35){\footnotesize $\Vert x_i-G_{y\rightarrow x}(G_{x\rightarrow y}(x_i))\Vert_p$}
		\put(-246,25){\footnotesize (a) DiscoGAN~\cite{kim2017learning}, CycleGAN~\cite{zhu2017unpaired}, }
		\put(-246,15){\footnotesize DualGAN~\cite{yi2017dualgan}, MUNIT~\cite{huang2018multimodal}, StarGAN~\cite{choi2018stargan}, }
		\put(-246,5){\footnotesize BicycleGAN~\cite{zhu2017toward}, DRIT++~\cite{lee2020drit++} $\dots$}
		\put(-84,35){\footnotesize $\Vert f(x_i)-f(G_{x\rightarrow y}(x_i))\Vert_p$}
		\put(-95,25){\footnotesize (b) DeePSiM~\cite{dosovitskiy2016generating}, StyleNet~\cite{ulyanov2017improved}, }
		\put(-95,15){\footnotesize PerceptualLoss~\cite{johnson2016perceptual}, SimGAN~\cite{shrivastava2017learning}, }
		\put(-95,5){\footnotesize ContextualLoss~\cite{mechrez2018contextual}, TSIT~\cite{jiang2020tsit} $\dots$}
		\put(47,35){\footnotesize $f(x_i) \propto f(G_{x\rightarrow y}(x_i))$}
		\put(38,25){\footnotesize (c) DistanceGAN~\cite{benaim2017one},}
		\put(38,15){\footnotesize TraVelGAN~\cite{amodio2019travelgan}, GcGAN~\cite{fu2019geometry}, }
		\put(38,5){\footnotesize CUT~\cite{park2020cut} $\dots$}
		\put(158,35){\footnotesize $d(S(x_i),S(G_{x\rightarrow y}(x_i)))$}
		\put(182,15){\footnotesize (d) Ours}
	\end{picture}
	
	\caption{\textbf{Comparison of unpaired I2I translation methods with various content losses}. (a) The cycle-consistency loss~\cite{kim2017learning,zhu2017unpaired,yi2017dualgan} in a two-sided framework. (b) Pixel-level image reconstruction loss~\cite{shrivastava2017learning} and feature-level matching loss~\cite{johnson2016perceptual}. (c) Various indirect relationships~\cite{benaim2017one,fu2019geometry} between the input and output. (d) Our spatially-correlative loss based on a learned spatially-correlative map. } 
	\label{fig:losses}	
\end{figure*}

\section{Introduction}

I2I translation refers to the task of modifying an input image to fit the \emph{style / appearance} of the target domain, while preserving the original \emph{content / structure } (as shown in Fig.~\ref{fig:example}: \emph{horse} $\rightarrow$ \emph{zebra}); learning to assess the \emph{content} and \emph{style} correctly is thus of central importance. While GANs~\cite{goodfellow2014generative} have the ability to generate images that adhere to the overall dataset distribution, it is still difficult to preserve scene structure during translation when image-conditional GANs are optimized with purely adversarial loss.

To mitigate the issue of scene structure discrepancies, a few loss functions for comparing the content between input and output images have been proposed, including (a) \emph{pixel-level} image reconstruction loss~\cite{isola2017image, shrivastava2017learning,chen2017photographic} and cycle-consistency loss~\cite{kim2017learning,zhu2017unpaired,yi2017dualgan}; (b) \emph{feature-level} perceptual loss~\cite{dosovitskiy2016generating,johnson2016perceptual} and PatchNCE loss~\cite{park2020cut}. However, these losses still have several limitations. First, pixel-level losses do \emph{not} explicitly decouple structure and appearance. Second, feature-level losses help but continue to conflate domain-specific structure and appearance attributes. Finally, most feature-level losses are calculated using a fixed ImageNet~\cite{deng2009imagenet} pre-trained network (\eg VGG16~\cite{simonyan2014very}), which will not correctly adapt to arbitrary domains.

In this work, we aim to design a \emph{domain-invariant} representation to precisely express scene structure, rather than using original pixels or features that couple both appearance and structure. To achieve this, we propose to revisit the idea of \emph{self-similarity}. Classically, low-level self-similarity has been used for matching~\cite{shechtman2007matching} and image segmentation~\cite{shi2000normcuts}, while feature-level self-similarity in deep learning manifests as self-attention maps~\cite{xu2015show}. We propose to go further, to advance an assumption that \emph{all} regions within same categories exhibit some form of self-similarity. For instance, while the horse and zebra in Fig.~\ref{fig:example} appear very different, there is obvious visual self-similarity in their own regions. We believe a network can learn deeper representations of self-similarities (beyond just visual ones) that can encode intact object shapes, even when there are variations in appearances within an object. Then through estimating such co-occurrence signals in self-similarity, we can explicitly represent the structure as multiple \emph{spatially-correlative} maps, visualized as heat maps in Fig.\ref{fig:example} (c) and (d). Based on this within-shape self-similarity, we propose then that \emph{a structure-preserving image translation will retain the patterns of self-similarity in both the source and translated images, even if appearances themselves change dramatically}. 

Our basic spatially-correlative map, called \emph{FSeSim}, is obtained by computing the \textbf{F}ixed \textbf{Se}lf-\textbf{Sim}ilarity of features extracted from a pre-trained network. While this basic version achieved comparative or even better results than state-of-the-art methods~\cite{zhu2017unpaired,fu2019geometry,park2020cut} on some tasks, the generality is limited because features extracted from an ImageNet pre-trained network are biased towards photorealistic imagery.

To obtain a more general spatially-correlative map, the \textbf{L}earned \textbf{Se}lf-\textbf{Sim}ilarity, called \emph{LSeSim}, is presented by using a form of contrastive loss, in which we explicitly encourage homologous structures to be closer, regardless of their appearances, and reciprocally dissociate dissimilar structures even they have similar appearances. To do this, the model learns a domain-invariant spatially-correlative map, where having the same scene structure leads to similar maps, even if the images are from different domains.

There are several advantages of using the proposed F/LSeSim loss: (a) In contrast to the existing losses that directly compare the loss on pixels \cite{zhu2017unpaired} or features \cite{johnson2016perceptual}, F/LSeSim captures the domain-invariant structure representation, regardless of the absolute pixel values; (b) Through contrastive learning, the LSeSim learns a spatially-correlative map for a specific image translation task, rather than features extracted from a fixed pre-trained network, as in \eg perceptual loss~\cite{johnson2016perceptual}, contextual loss \cite{mechrez2018contextual}; (c) The translation model is more efficient and faster than the widely used cycle-consistency architectures, because our F/LSeSim explicitly encodes the structure, bypassing the expensive multi-cycle looping; (d) As we show in Fig.~\ref{fig:error_map}, our F/LSeSim correctly measures the structural distance even when the two images are in completely different domains; (e) Finally, our F/LSeSim can easily be integrated into various frameworks. In our experiments, we directly used the generator and discriminator architectures of CycleGAN~\cite{zhu2017unpaired}, MUNIT~\cite{huang2018multimodal} and StyleGAN~\cite{karras2019style,karras2020analyzing} for extensive I2I translation tasks. The experimental results show that our model outperformed the existing both one-sided translation methods~\cite{benaim2017one,amodio2019travelgan,fu2019geometry,park2020cut} and two-sided translation methods~\cite{zhu2017unpaired,yi2017dualgan,huang2018multimodal}.

\section{Related Work}

Existing unpaired I2I translation either use  cycle-consistency loss in a two-sided framework~\cite{kim2017learning,zhu2017unpaired,yi2017dualgan}, or other forms of pixel-level and feature-level losses in a one-sided framework~\cite{benaim2017one,amodio2019travelgan,fu2019geometry} for preserving content (Fig.~\ref{fig:losses}).

\vspace{-0.2cm}\paragraph{Two-Sided Unsupervised Image Translation.} \emph{Cycle-consistency} has became a de facto loss in most works, whether the cycles occur in the image domain~\cite{zhu2017unpaired,kim2017learning,yi2017dualgan,choi2018stargan,hoffman2018cycada,lee2018diverse}, or in latent space~\cite{zhu2017toward,huang2018multimodal,lee2020drit++}. However, without explicit constraints, the content in a translated image can be easily distorted~\cite{lee2018diverse}. Furthermore, the cycle-based methods require auxiliary generators and discriminators for the reverse mapping.

\vspace{-0.2cm}\paragraph{One-Sided Unsupervised Image Translation.} To avoid cycle-consistency artifacts, 
DistanceGAN~\cite{benaim2017one} and GcGAN~\cite{fu2019geometry} pre-define an implicit distance in a one-sided framework. In contrast, the feature-level losses~\cite{johnson2016perceptual,mechrez2018contextual} evaluate the content distance in a deep feature space, which have been applied in both style transfer~\cite{gatys2016image,johnson2016perceptual,mechrez2018contextual,zhang2018unreasonable} and image translation~\cite{chen2017photographic,wang2018high,park2019semantic,jiang2020tsit}. However, the underlying assumption that high-level semantic information is solely determined in feature space does not always hold. Furthermore, these features are extracted from a fixed pre-trained network (\eg VGG16~\cite{simonyan2014very}). While the latest CUT~\cite{park2020cut} learns a PatchNCE loss for a specific task, the distance used is directly computed from extracted features, and will still be affected by domain-specific peculiarities. 

\vspace{-0.2cm}\paragraph{Contrastive Representation Learning.} Driven by the potential of discriminative thought, a series of self-supervised methods~\cite{hjelm2018learning,wu2018unsupervised,oord2018representation,bachman2019learning,henaff2019data,chen2020simple,he2020momentum} have emerged in recent years. These self-supervised methods learn robust features by associating ``positive'' pairs and dissociating ``negative'' pairs. CUT~\cite{park2020cut} first introduced contrastive learning for unsupervised I2I translation. While we utilize a patch-wise contrastive loss within an image in a similar manner to CUT, we propose a systematic way to learn a structure map that excludes appearance attributes. 

\section{Methods}\label{methods}

As shown in Fig.~\ref{fig:losses}, given a collection of images $\mathcal{X}\subset\mathbb{R}^{H\times W\times C}$ from a particular domain (\eg horse), our main goal is to learn a model $\Phi$ that receives the image $x\in \mathcal{X}$ as input and transfers it into the target domain $\mathcal{Y}\subset\mathbb{R}^{H\times W\times C}$ (\eg zebra), in a manner that retains the original scene structure but converts the appearance appropriately. 

Here, we focus on designing a loss function that measures the structural similarity between the input image $x$ and the translated image $\hat{y}=\Phi(x)$.
However, unlike most existing approaches that directly attempt to evaluate the structural similarity between input and translated images at some deep feature level, we will instead compute the \emph{self-similarity} of deep features \emph{within each image}, and then \emph{compare the self-similarity patterns} between the images.

In subsequent sections, we investigate two losses, \emph{fixed self-similarity (FSeSim)} and \emph{learned self-similarity (LSeSim)}. In the first instance, we directly compare the self-similarity patterns of features extracted from a fixed pre-trained network (\eg VGG16~\cite{simonyan2014very}). In the second instance, we additionally introduce a structure representation model that learns to correctly compare the self-similarity patterns, in which we use the contrastive infoNCE loss~\cite{oord2018representation} to learn such a network without label supervision. 

\begin{figure}[tb!]
	\centering
	\includegraphics[width=\linewidth]{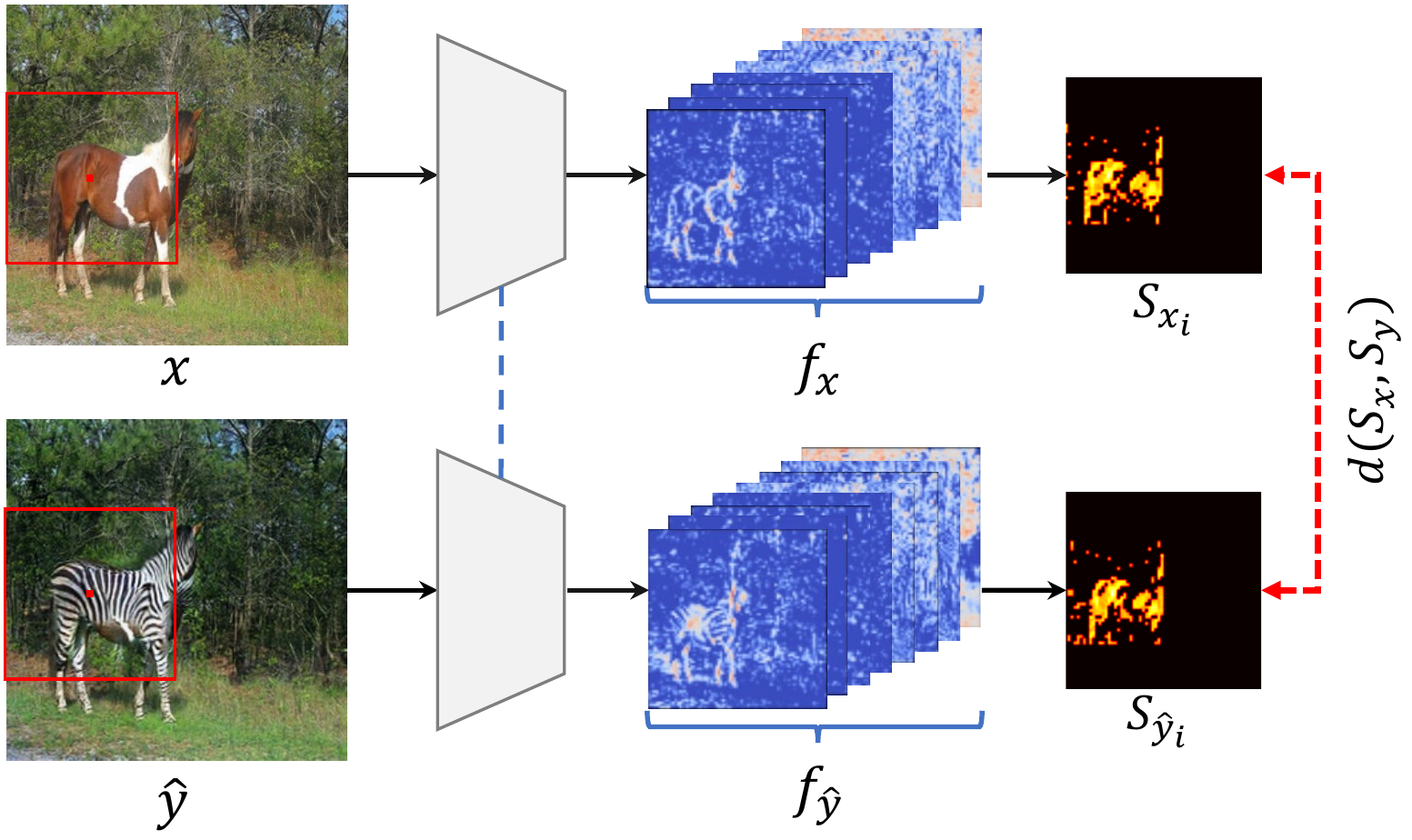}
	\begin{picture}(0,0)
	\put(-32,78){\rotatebox{90}{\footnotesize shared}}
	\put(-100,7){\footnotesize Image}
	\put(-47,15){\footnotesize Feature}
	\put(-49,7){\footnotesize extractor}
	\put(5,7){\footnotesize Features}
	\put(55,7){\footnotesize Similarity map}
	\end{picture}
	\caption{\textbf{An example of computing spatially-correlative loss from self-similarity maps.} Image $x$ and corresponding translated image $\hat{y}$ are first fed into the feature extractor. We then compute the local self-similarity for each query point. Here, we show one example for the red query point. }
	\label{fig:framework}
\end{figure}
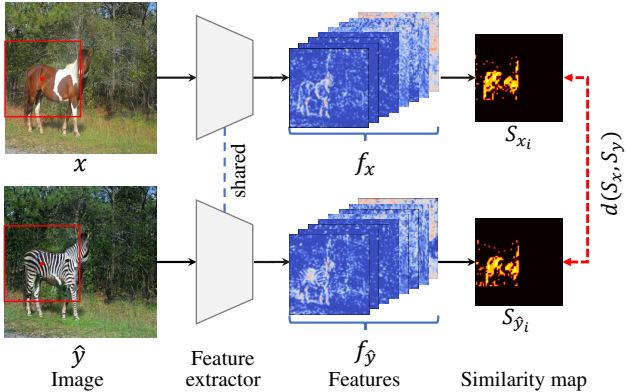 

\subsection{Fixed Self-Similarity (FSeSim)}\label{sec:FSeSim}

We first describe our fixed spatially-correlative loss.
Given an image $x$ in one domain and its corresponding translated image $\hat{y}$ in another, we extract the features $f_x$ and $f_{\hat{y}}$ using a simple network (\eg VGG16~\cite{simonyan2014very}). Instead of directly computing the feature distance $\Vert f_x-f_{\hat{y}} \Vert_p $, we compute the self-similarity in the form of a map. We call this a \emph{spatially-correlative map}, formally:
\begin{equation}
	S_{x_i} = (f_{x_i})^T(f_{x_*})
\end{equation}
where $f_{x_i}^T\in\mathbb{R}^{1\times C}$ is the feature of a query point $x_i$ with $C$ channels, $f_{x_*}\in\mathbb{R}^{C\times N_p}$ contains corresponding features in a patch of $N_p$ points, and $S_{x_i}\in\mathbb{R}^{1\times N_p}$ captures the feature spatial correlation between the query point and other points in the patch. We show one query example in Fig.~\ref{fig:framework}, where the spatially-correlative map for the query patch is visualized as a heat map. Note that unlike the original features that would still encode domain-specific attributes such as color, lighting and texture, the self-similarity map only captures the spatially-correlative relationships. 

Next we represent the structure of the whole image as a collection of multiple spatially-correlative maps $S_x=[S_{x_1};S_{x_2};\dots;S_{x_s}]\in\mathbb{R}^{N_s\times N_p}$, where $N_s$ is the numbers of sampled patches. This is a semi-sparse representation, but is more computationally efficient. We then compare the multiple structure similarity maps between the input $x$ and the translated image $\hat{y}$, as follows:
\begin{equation}\label{eq:FSeSimL}
	\mathcal{L}_{s} = d(S_x, S_{\hat{y}})
\end{equation} 
where $S_{\hat{y}}$ are corresponding spatially-correlative maps in the target domain. Here, we consider two forms for $d(\cdot)$, the $L_1$ distance $\Vert S_x-S_{\hat{y}} \Vert_1$ and the cosine distance $\Vert 1-\cos(S_x, S_{\hat{y}})\Vert$. The former term strongly encourages the spatial similarity to be consistent at all points in a patch, while the latter term supports pattern correlation without concern for differences in magnitude. 

\begin{figure}[tb!]
	\centering
	\includegraphics[width=\linewidth]{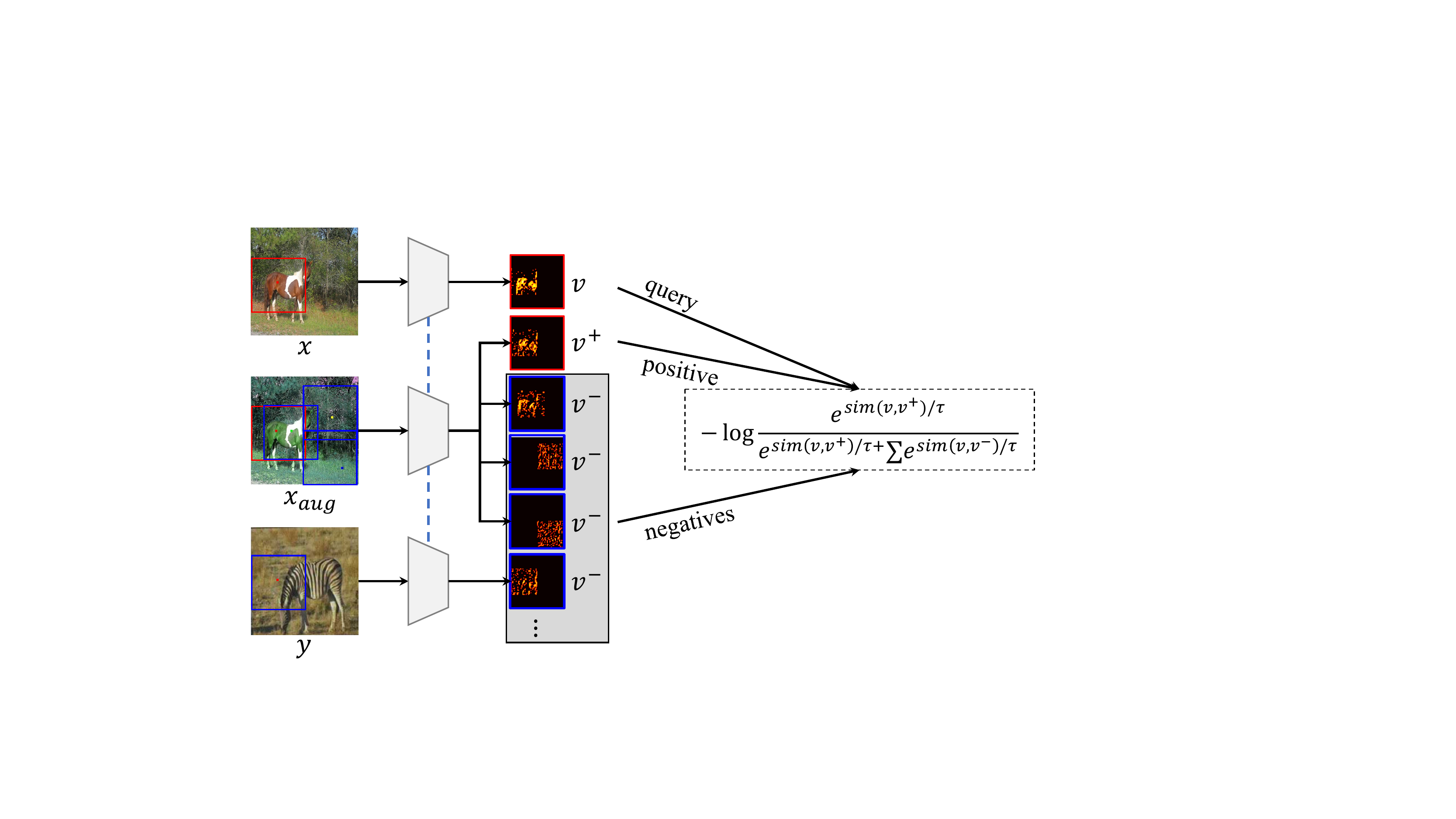}
	\begin{picture}(0,0)
	\put(-62,95){\rotatebox{90}{\footnotesize shared}}
	\put(-62,50){\rotatebox{90}{\footnotesize shared}}
	\put(-112,4){\footnotesize Image}
	\put(-76,13){\footnotesize Feature}
	\put(-79,4){\footnotesize extractor}
	\put(-45,13){\footnotesize Sample \textcolor{red}{positive}}
	\put(-43,4){\footnotesize + \textcolor{blue}{K negatives}}
	\put(42,4){\footnotesize Contrastive loss}
	\end{picture}
	\caption{\textbf{Patchwise contrastive learning for the learned self-similarity.} Three images are fed into the feature extractor, in which two images, $x$ and $x_{aug}$, are homologous with the same structure but varied appearances, and $y$ is another randomly sampled image. For each query patch in $x$, the ``positive'' sample is the corresponding patch in $x_{aug}$, and all other patches are considered as ``negative'' samples.}
	\label{fig:learned_attn}
\end{figure}
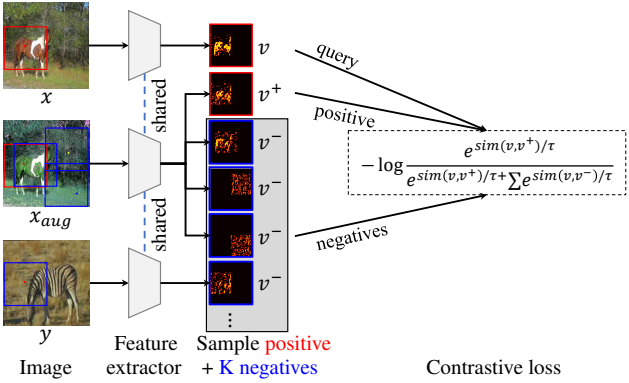 

\subsection{Learned Self-Similarity (LSeSim)}\label{sec:LSeSim}

Although our FSeSim provides strong supervision for structure consistency, it does \emph{not} explicitly learn a structure representation for a specific translation task. As opposed to existing feature-level losses~\cite{johnson2016perceptual,mechrez2018contextual} that only utilize the features from a fixed pre-trained network, we propose to additionally learn a structure representation network for each task that expresses the learned self-similarity, or LSeSim.

In order to learn such a model \emph{without supervision}, we consider the self-supervised contrastive learning that associates similar features, while simultaneously dissociates different features. Following PatchNCE~\cite{park2020cut}, we build our contrastive loss at patch level, except here the pairs for comparison are our spatially-correlative maps, rather than the original features in existing works~\cite{hjelm2018learning,chen2020simple,he2020momentum,park2020cut}. To help generate pairs of similar patch features for self-supervised learning, we create augmented images by applying structure-preserving transformations. 

Formally, let $\boldsymbol{v}=S_{x_i}\in\mathbb{R}^{1\times N_p}$ denotes the spatially-correlative map of the ``query'' patch. Let $\boldsymbol{v}^+=S_{\hat{x}_i}\in\mathbb{R}^{1\times N_p}$ and $\boldsymbol{v}^-\in\mathbb{R}^{K \times N_p}$ be ``positive'' and ``negative'' patch samples, respectively. The query patch is positively paired with a patch in the same position $i$ within an augmented image $x_{aug}$, and negatively paired to patches sampled from other positions in $x_{aug}$, or patches from other images $y$. The number of negative patches used is $K=255$.

Our LSeSim design is illustrated in Fig.~\ref{fig:learned_attn}. The contrastive loss is given by:
\begin{equation}
	\mathcal{L}_c = -\log\frac{e^{sim(\boldsymbol{v},\boldsymbol{v}^+)/\tau}}
	{e^{sim(\boldsymbol{v},\boldsymbol{v}^+)/\tau}+\sum_{k=1}^{K}e^{sim(\boldsymbol{v},\boldsymbol{v}_k^-)/\tau}}
\end{equation}
where $sim(\boldsymbol{v},\boldsymbol{v}^+)=\boldsymbol{v}^T\boldsymbol{v}^+/\Vert\boldsymbol{v}\Vert\Vert\boldsymbol{v}^+\Vert$ is the cosine similarity between two spatially-correlative maps, and $\tau$ is a temperature parameter. To minimize this loss, our network encourages the corresponding patches with the same structure to be close even they have very different visual appearances, which fits in with the goal of image translation. Note that, this contrastive loss is only used for optimizing the structure representation network. The spatially-correlative loss for the generator is always the loss in eq.\ (\ref{eq:FSeSimL}).

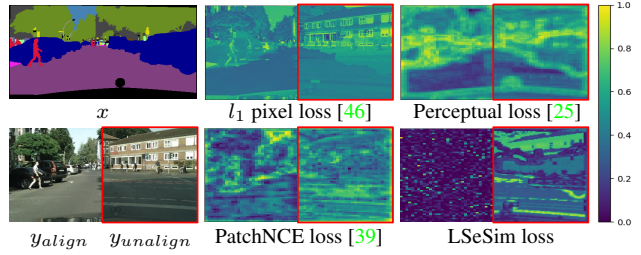
\begin{figure}[tb!]
	\centering
	\includegraphics[width=\linewidth]{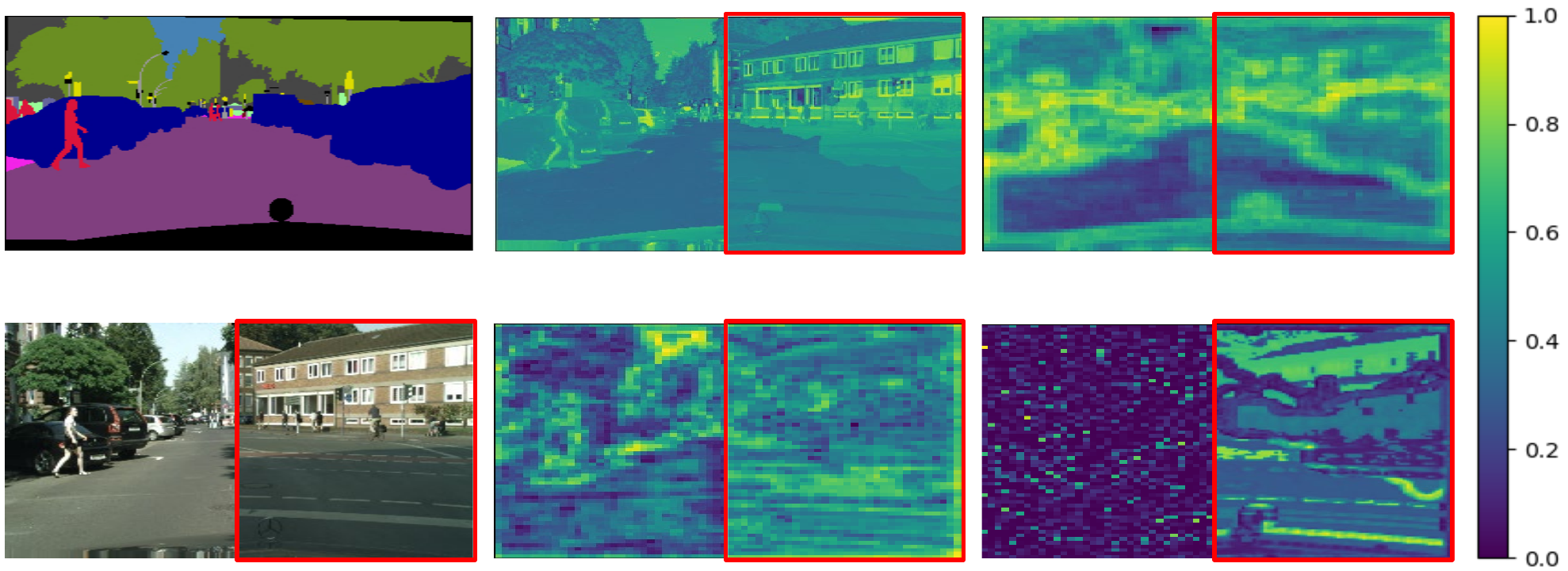}
	\begin{picture}(0,0)
		\put(-85,53){\footnotesize $x$}
		\put(-35,53){\footnotesize $l_1$ pixel loss~\cite{shrivastava2017learning}}
		\put(35,53){\footnotesize Perceptual loss~\cite{johnson2016perceptual}}
		\put(-110,6){\footnotesize $y_{align}$}
		\put(-80,6){\footnotesize $y_{unalign}$}
		\put(-40,6){\footnotesize PatchNCE loss~\cite{park2020cut}}
		\put(48,6){\footnotesize LSeSim loss}
	\end{picture}
	\caption{\textbf{Error map visualization.} Our LSeSim has small errors on the left where ground truth paired data is provided, while having large errors on the right for unpaired data.}
	\label{fig:error_map}
\end{figure}

\begin{figure*}[tb!]
	\centering
	\setlength{\abovecaptionskip}{0.cm}
	\setlength{\belowcaptionskip}{-0.cm}
	\includegraphics[width=\linewidth,height=0.09\textheight]{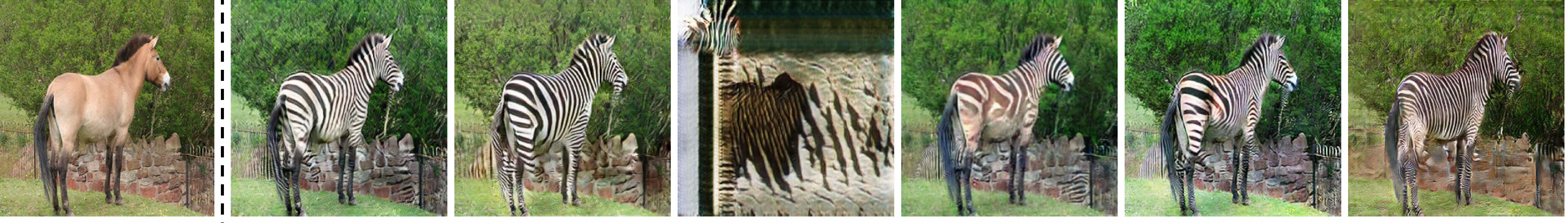}
	\begin{picture}(0,0)
		\put(-222,5){\footnotesize Input}
		\put(-152,5){\footnotesize LSeSim}
		\put(-81,5){\footnotesize FSeSim }
		\put(-30,5){\footnotesize GAN loss only~\cite{zhu2017unpaired}}
		\put(49,5){\footnotesize Cycle loss~\cite{zhu2017unpaired}}
		\put(111,5){\footnotesize Perceptual loss~\cite{johnson2016perceptual}}
		\put(182,5){\footnotesize PatchNCE loss~\cite{park2020cut}}
	\end{picture}
	\caption{\textbf{Comparing results under different content losses}. All results are reported following the same setting of CycleGAN~\cite{zhu2017unpaired}, except using different content losses. Our model generates much better visual results with only loss modification. }
	\label{fig:loss_comp}	
\end{figure*}
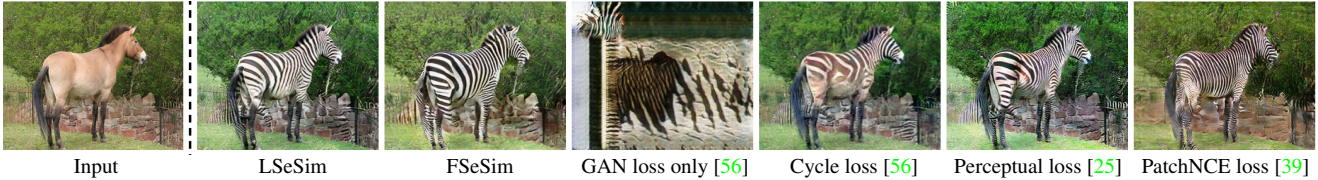

\subsection{Full Objective}
Overall, we train the networks by jointly minimizing the following losses:
\begin{equation}\label{eq:all}
	\begin{split}
	\mathcal{L}_D &= -\mathbb{E}_{y\sim p_{d}}[\log D(y)]-\mathbb{E}_{\hat{y}\sim p_{g}}[\log(1-D(\hat{y}))] \\
	\mathcal{L}_S &= \mathcal{L}_c \\
	\mathcal{L}_G &= \mathbb{E}_{\hat{y}\sim p_{g}}[\log(1-D(\hat{y}))] + \lambda d(S_x, S_{\hat{y}}) \\
	\end{split}
\end{equation}
where $\mathcal{L}_D$ is the adversarial loss for the discriminator $D(\cdot)$, $\hat{y}$ is the translated image, and $\mathcal{L}_S$ is the contrastive loss for the structure representation network $f(\cdot)$. $\mathcal{L}_G$ is the loss for the generation (translation) network $G(\cdot)$, which consists of the style loss term and the structure loss term. $\lambda$ is a hyper-parameter to trade off between style and content.

\subsection{Analysis}\label{sec:analysis}

Readers may wonder why the proposed F/LSeSim losses would perform better than existing feature-level losses~\cite{johnson2016perceptual,mechrez2018contextual,park2020cut}. An intuitive interpretation is that \emph{self-similarity deals only with spatial relationships of co-occurring signals, rather than their original absolute values}.

To provide further clarity, we consider a scenario where given a semantic map $x$ (Fig.~\ref{fig:error_map}), the task is to translate it to a photorealistic image $y$. We consider an ideal result (the paired ground truth $y_{align}$) and a wrong result (another image $y_{unalign}$), respectively. Under such a setting, a good structure loss should penalize the wrong result, while supporting the ideal result. To visualize the error maps, for each corresponding pair of query patches in $x$ and $y$ we computed the error at that patch location for different losses. As can be seen, pixel-level loss~\cite{shrivastava2017learning} is naturally unsuitable when there are large domain gaps, and while Perceptual loss~\cite{johnson2016perceptual} will report significant errors for both aligned and unaligned results. PatchNCE~\cite{park2020cut} mitigates the problem by calculating the cosine distance of features, but it can be seen the loss map still retains high errors in many regions within the aligned result, due to extracted features consisting of appearance attributes, such as color and texture.

In contrast, appearance attributes are ignored in LSeSim by representing scene structure as a spatially-correlative map. Fig.~\ref{fig:error_map} shows that our LSeSim leads to low errors for the aligned image (left), even when they are in quite different domains, but large errors for the non-aligned image (right). Even for $y_{unalign}$, LSeSim differentiates between related structures (\eg roads) and unrelated structures (trees vs windows), with lower errors for the former. Hence LSeSim can better help preserve scene structure even across large domain gaps. 

In Fig.~\ref{fig:loss_comp}, we report a qualitative comparison of various losses that be applied to a same translation network architecture. All methods following the setting in CycleGAN~\cite{zhu2017toward}, except that the content loss is changed. Cycle-consistency is achieved using the auxiliary generator and discriminator, and all other methods are one-sided translation. We find that our method produces results with much better visual quality.

\vspace{-0.2cm}\paragraph{Discussion.} Similar to conventional feature-level losses \cite{johnson2016perceptual,mechrez2018contextual}, our F/LSeSim is computed in a deep feature space. However, we represent the structure as multiple spatially-correlative maps. So rather than directly at feature level which is not free from domain-specific attributes, our comparison is done at a more abstract level that is intended to transcend domain specificity.

While attention maps have been used in previous image translation works~\cite{chen2018attention,alami2018unsupervised}, it is fundamentally different from our F/LSeSim in concept --- their attention maps effectively function as saliency maps to guide the translation, but content preservation is primarily still dependent on cycle-consistency loss. In our case, the multiple spatially-correlative maps are used to encode and determine invariance in scene structure. Our F/LSeSim also differs from the content loss used in~\cite{kolkin2019style}, in which the self-similarity was calculated at random positions without a clear purpose. Our F/LSeSim is on the other hand organized at a local patch level to explicitly represent the scene structure. As shown in \S~\ref{res:multi}, our local structure representation is better than just using random spatial relationships. Furthermore, our LSeSim is a metric learned from the infoNCE loss, which generalizes well robustly on various tasks. While PatchNCE loss~\cite{park2020cut} can also learn feature similarity using contrastive loss, it directly compares features in two domains. 

\section{Experiments}

To demonstrate the generality of our method, we instantiated F/LSeSim in multiple frameworks on various I2I translation tasks, including \emph{single-modal} , \emph{multi-modal}, and even \emph{single-image} translation. For each task, we used a suitable baseline architecture, but replaced their content losses with our F/LSeSim loss. In addition, we are only interested in scenarios where scene structure is preserved during the translation~\cite{isola2017image,zhu2017unpaired,zhu2017toward}, rather investigating translations incorporating shape modification \cite{choi2018stargan,choi2020stargan,nizan2020breaking,Kim20DST,baek2020rethinking}.  

\subsection{\emph{Single-Modal} Unpaired Image Translation}\label{res:singlemodal}

We first evaluated our loss on the classical single-modal unpaired I2I translation task. 
\vspace{-0.2cm}\paragraph{Implementation details. } In this task, we chose CycleGAN \cite{zhu2017unpaired} as the reference architecture, but only used half of their pipeline and replaced the cycle-consistency loss with our F/LSeSim loss. Specifically, we used the Resnet-based generator with PatchGAN discriminator~\cite{isola2017image}. Details can be found on their \href{https://github.com/junyanz/pytorch-CycleGAN-and-pix2pix}{website}.

Our FSeSim is based on the ImageNet-pretrained VGG16 \cite{simonyan2014very}, where we used features from layers \texttt{relu}3\_1 and \texttt{relu}4\_1. While the LSeSim employs the same structure as FSeSim, the weights are not fixed and additionally two convolution layers, implemented as $1\times1$ kernels, are included to select better features. As for the selection of patches to build the contrastive loss, we found that random sampling the patch locations performed much better than uniform sampling on a grid, leading to better convergence when training the structure representation network. We set $\lambda=10$ in FSeSim and $\tau=0.07$ in LSeSim.

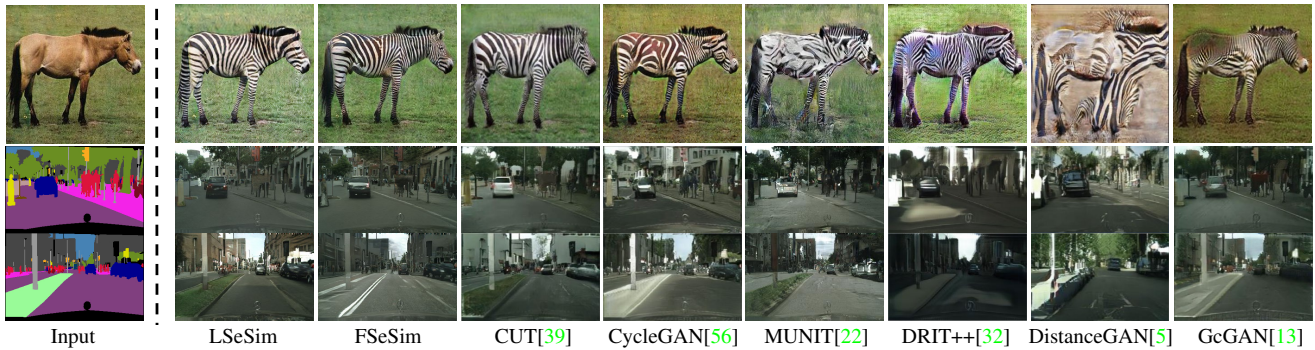
\begin{figure*}[tb!]
	\centering
	\setlength{\abovecaptionskip}{0.cm}
	\setlength{\belowcaptionskip}{-0.cm}
	\includegraphics[width=\linewidth]{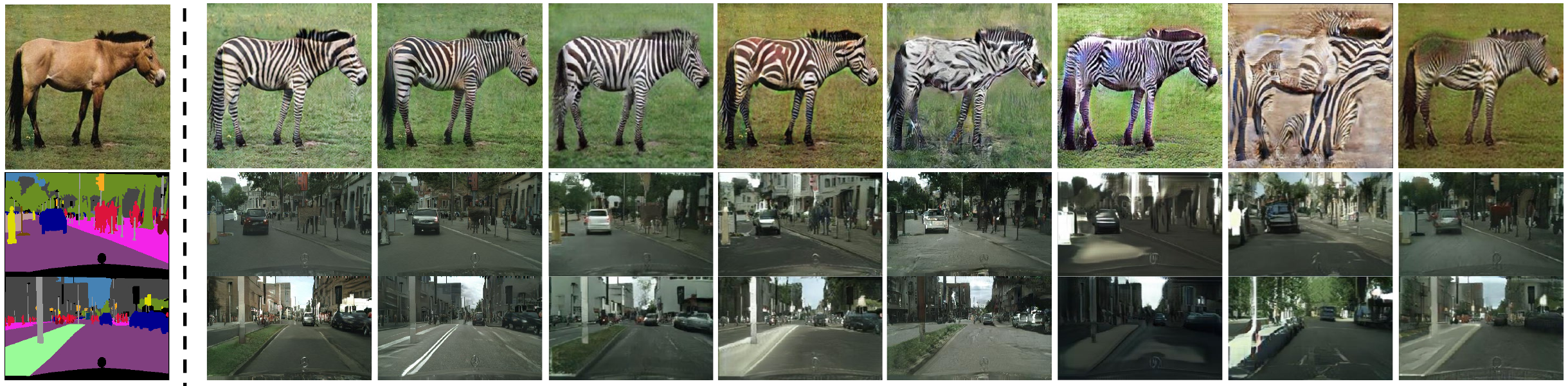}
	\begin{picture}(0,0)
		\put(-230,5){\footnotesize Input}
		\put(-170,5){\footnotesize LSeSim}
		\put(-115,5){\footnotesize FSeSim}
		\put(-62,5){\footnotesize CUT\cite{park2020cut}}
		\put(-19,5){\footnotesize CycleGAN\cite{zhu2017unpaired}}
		\put(40,5){\footnotesize MUNIT\cite{huang2018multimodal}}
		\put(92,5){\footnotesize DRIT++\cite{lee2020drit++}}
		\put(140,5){\footnotesize DistanceGAN\cite{benaim2017one}}
		\put(204,5){\footnotesize GcGAN\cite{fu2019geometry}}
	\end{picture}
	\caption{\textbf{Qualitative comparison on single-modal image translation.} Here, we show results for \emph{horse}$\rightarrow$\emph{zebra} and \emph{label}$\rightarrow$ \emph{image}.}
	\label{fig:results_sin_mod}	
\end{figure*}

\begin{table}[tb!]
	\centering
	\renewcommand{\arraystretch}{1.1}
	\begin{tabular}{@{}lccccc@{}}
		\hlineB{3.5}
		\multirow{2}{*}{\textbf{Method}} & \multicolumn{2}{c}{\textbf{Cityscapes}}&& \multicolumn{2}{c}{\textbf{Horse$\rightarrow$Zebra}}\\
		\cline{2-3}\cline{5-6}
		& pixAcc$\uparrow$& FID$\downarrow$ && FID$\downarrow$ & Mem$\downarrow$\\
		\hlineB{2}
		CycleGAN~\cite{zhu2017unpaired} & 57.2 & 76.3 && 77.2 & 4.81 \\
		MUNIT~\cite{huang2018multimodal}  & 58.4 & 91.4 && 98.0 & 9.43\\
		DRIT++~\cite{lee2020drit++} & 60.3 & 96.2 && 88.5 & 11.2\\
		\cdashline{1-6}
		Distance~\cite{benaim2017one} & 47.2 & 75.9 && 67.2& 2.72\\
		GcGAN~\cite{fu2019geometry} & 65.5 & 57.4 && 86.7 & 4.68 \\
		CUT~\cite{park2020cut} & 68.8 & 56.4 && 45.5 & 3.33 \\
		\cdashline{1-6}
		FSeSim & 69.4 & 53.6 && 40.4 & \textbf{2.65} \\
		LSeSim & \textbf{73.2} & \textbf{49.7} && \textbf{38.0} & 2.92 \\
		\hlineB{2}
	\end{tabular}
	\caption{\textbf{Quantitative comparison on single-modal image translation}. FID~\cite{heusel2017gans} measures the distance between distributions of generated images and real images. ``Mem'' denotes the memory cost during training.}
	\label{table:single_mod}
\end{table}

\vspace{-0.2cm}\paragraph{Metrics.} Our evaluation protocols are adopted from previous work~\cite{heusel2017gans,park2019semantic,park2020cut}. We first used the popular Fr\'echet Inception Distance (FID)~\cite{heusel2017gans} \footnote{As claimed in StyleGANv2~\cite{karras2020analyzing}, ImageNet-pretrained classifiers tend to evaluate the distribution on texture than shape, while humans focus on shape. The best FID score does \emph{not} ensure the best image quality for translated images. As such, for a fair comparison, we reported the best FID score from all trained epochs for all methods, rather than the score in the last epoch. } to assess the visual quality of generated images by comparing the distance between distributions of generated and real images in a deep feature domain. For \emph{semantic image synthesis}, we further applied semantic segmentation to the generated images to estimate how well the predicted masks match the ground truth segmentation masks as in~\cite{chen2017photographic,wang2018high,park2019semantic,park2020cut}. Following~\cite{park2020cut,jiang2020tsit}, we used the pre-trained DRN~\cite{yu2017dilated}.

\vspace{-0.2cm}\paragraph{Results.}  In Table~\ref{table:single_mod}, we reported either published results or our reproductions with publicly-available code, choosing the better. Our simple, inexpensive losses substantially outperformed state-of-the-art methods, including two-sided frameworks with multiple cycle-consistency losses~\cite{zhu2017toward,huang2018multimodal,lee2020drit++}, and one-sided frameworks using self-distance~\cite{benaim2017one}, geometry consistency~\cite{fu2019geometry} and contrastive loss~\cite{park2020cut}. 

When compared to CycleGAN~\cite{zhu2017unpaired} and CUT~\cite{park2020cut}, although we used the same settings for the generator and discriminator, our method led to significant improvement. Unlike CUT~\cite{park2020cut} that depends on an identity pass for good performance, our results were achieved by training with only one pass using F/LSeSim and GAN losses. This suggests that once we explicitly decouple scene structure and appearance, it is easier for the model to modify the visual appearance correctly. As our model belongs to one-sided image translation that does \emph{not} require additional generators and discriminators, our model is also memory-efficient.  

Qualitative results are shown in Figs.~\ref{fig:loss_comp} and ~\ref{fig:results_sin_mod}. In Fig.~\ref{fig:loss_comp}, despite keeping the same settings and only comparing different content losses, our method translated the zebra appearance more cleanly. We also compared results using the same examples as \cite{park2020cut} in Fig.~\ref{fig:results_sin_mod}, where our method achieved better visual results.

\begin{figure*}[tb!]
	\centering
	\setlength{\abovecaptionskip}{0.cm}
	\setlength{\belowcaptionskip}{-0.cm}
	\includegraphics[width=\linewidth]{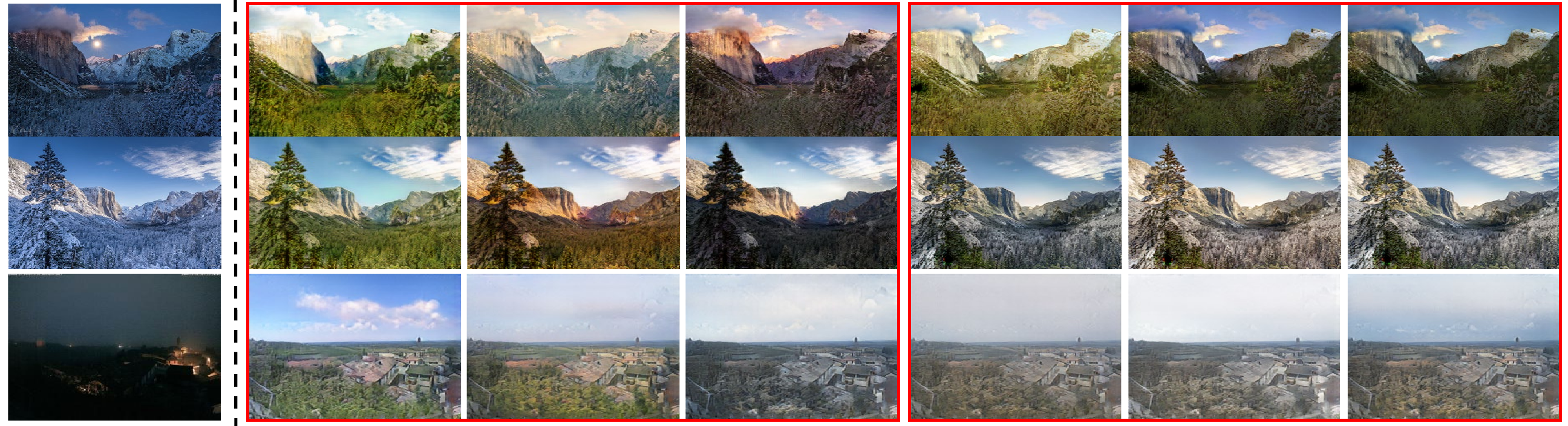}
	\begin{picture}(0,0)
		\put(-220,5){\footnotesize Input}
		\put(-80,5){\footnotesize LSeSim}
		\put(126,5){\footnotesize MUNIT\cite{huang2018multimodal}}
	\end{picture}
	
	\caption{\textbf{Qualitative comparison on multi-modal image translation}. Here, we show the examples of \emph{winter}$\rightarrow$\emph{summer} and \emph{night}$\rightarrow$\emph{day}.} 
	\label{fig:results_mul_mod}	
\end{figure*}
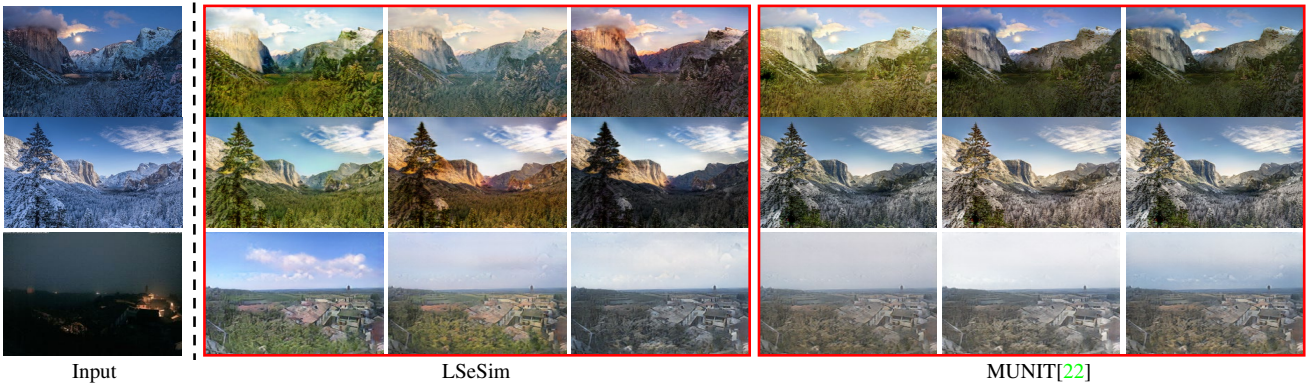

\begin{table*}[tb!]
	\centering
	\renewcommand{\arraystretch}{1.1}
	\begin{tabular}{@{}lccccccc@{}}
		\hlineB{3.5}
		\multirow{2}{*}{\textbf{Method}} & \multicolumn{3}{c}{\textbf{Winter$\rightarrow$Summer}}&& \multicolumn{3}{c}{\textbf{Night$\rightarrow$Day}}\\
		\cline{2-4}\cline{6-8}
		& LPIPS $\uparrow$& FID $\downarrow$ & D \& C $\uparrow$ && LPIPS $\uparrow$ & FID $\downarrow$ & D \& C $\uparrow$\\
		\hlineB{2}
		Real images & 0.770 & 44.1 & 0.997 / 0.986 && 0.684& 146.1 & 0.977 / 0.962 \\
		\cdashline{1-8}
		BicycleGAN~\cite{zhu2017toward} & \textbf{0.285} & 99.2 $\pm$ 3.2 & \textbf{0.536} / 0.667 && \textbf{0.349} & 290.9 $\pm$ 6.5 & \textbf{0.375} / 0.515 \\
		MUNIT~\cite{huang2018multimodal} & 0.160 & 97.4 $\pm$ 2.2 & 0.439 / 0.707&& 0.152 & 267.1 $\pm$ 2.7 & 0.271 / 0.548 \\
		DRIT++~\cite{lee2020drit++} & 0.186 & 93.1 $\pm$ 2.0 & 0.494 / 0.753 && 0.167 & 258.5 $\pm$ 2.3 & 0.298 / 0.631 \\
		\cdashline{1-8}
		\textbf{FSeSim} & 0.216 & 90.5 $\pm$ 1.9 & 0.501 / 0.779 && 0.203 & 234.3 $\pm$ 2.8 & 0.332 / 0.638 \\
		\textbf{LSeSim} & 0.232 & \textbf{89.4 $\pm$ 1.9} & 0.516 / \textbf{0.793} && 0.215 & \textbf{224.9 $\pm$ 2.0}  & 0.347 / \textbf{0.652} \\
		\hlineB{2}
	\end{tabular}
	\caption{\textbf{Quantitative evaluation on multi-modal image translation task}. LPIPS distance~\cite{zhang2018unreasonable} measures the diversity of generated images by comparing the features of two images, while (D\&C)~\cite{ferjad2020icml} evaluates the diversity and fidelity by matching whole features in the generated and real datasets. }
	\label{table:multi_modal}
\end{table*}

\begin{table*}[tb!]
	\centering
	\renewcommand{\arraystretch}{1.1}
	\begin{tabular}{@{}llcccccc@{}}
		\hlineB{3.5}
		& \multirow{2}{*}{\textbf{Configuration}} & \multicolumn{2}{c}{\textbf{Horse $\rightarrow$ Zebra}} && \multicolumn{3}{c}{\textbf{Night $\rightarrow$ Day}} \\
		\cmidrule{3-4}\cmidrule{6-8}
		& & FID $\downarrow$ & Mem(GB) $\downarrow$ && FID $\downarrow$ & LPIPS $\uparrow$ & D \& C $\uparrow$ \\
		\hlineB{2}
		A & STROTSS~\cite{kolkin2019style} (random SeSim) & 70.1 & 2.68 && 262.7 $\pm$ 3.6  & 0.162 & 0.289 / 0.554 \\
		B & Baseline (global SeSim on single layer) & 53.7 & 2.97 && \textbf{173.2 $\pm$ 2.2} & 0.168 &  0.303 / \textbf{0.664} \\
		\cdashline{1-8}
		C & (B): Global $\rightarrow$ Patch ($32\times32$) & 45.8 & \textbf{2.61} && 231.3 $\pm$ 2.5 & 0.181 &  0.317 / 0.634 \\
		D & (C): Single $\rightarrow$ Multi (\texttt{relu3}\_1, \texttt{relu}4\_1) & 43.3 & 2.65 && 229.4 $\pm$ 2.1 & 0.177 &  0.311 / 0.646 \\
		E & (D): l1loss $\rightarrow$ 1 - $consine$& 40.4 & 2.65 && 234.3 $\pm$ 2.8 & 0.203 & 0.332 / 0.638  \\
		\cdashline{1-8}
		F & Ours LSeSim & \textbf{38.0} & 2.92 && 224.9 $\pm$ 2.0 &  \textbf{0.215} & \textbf{0.347} / 0.652 \\
		\hlineB{2}
	\end{tabular}
	\caption{\textbf{Ablation study on both single- and multi-modal image translation}. Refer to ablation experiments in main text for details.}
	\label{table:ablation}
\end{table*}

\subsection{\emph{Multi-Modal} Unpaired Image Translation}\label{res:multi}

Our F/LSeSim is also naturally suited for multi-modal image translation, since the use of our spatial-correlative maps imposes only structural consistency and not appearance constraints.
We performed a thorough comparison of F/LSeSim to state-of-the-art methods, along with comprehensive ablation experiments. 

\paragraph{Implementation details.} Our multi-modal setting is based on MUNIT~\cite{huang2018multimodal,lee2020drit++}, except our model uses only one generator and one discriminator of MUNIT~\cite{huang2018multimodal} without requiring the auxiliary generators and discriminators for multiple cycle training. Specifically, we used the Resnet-based generator with Instance Normalization (IN)~\cite{ulyanov2017improved} in the encoder and Adaptive Instance Normalization (AdaIN)~\cite{huang2017arbitrary,karras2019style} in the decoder, plus multi-scale discriminators~\cite{wang2018high}. The details of the architecture can be found on their \href{https://github.com/NVlabs/MUNIT}{website}. The F/LSeSim used is identical to that used in \S~\ref{res:singlemodal}. 

\paragraph{Metrics.} Besides using FID to measure quality, we also used the average LPIPS distance~\cite{zhang2018unreasonable} to evaluate the diversity of generated results. The LPIPS distance is calculated by comparing the features of two images. Following~\cite{huang2018multimodal,zhu2017toward}, we computed the distances between 1900 pairs, sampling 100 images 19 times. We also report the latest metrics of Density and Coverage (D\&C)~\cite{ferjad2020icml}, which separately evaluate the diversity and fidelity of generated results. Likewise, we used the 1900 sampled pairs to compute D\&C scores. Higher scores here indicate larger diversity and better coverage to the ground-truth domain, respectively. 

\paragraph{Results.} We compared our F/LSeSim to state-of-the-art methods in multi-modal image translation in Table~\ref{table:multi_modal}. Our method outperformed the two baselines, MUNIT~\cite{huang2018multimodal} and DRIT++~\cite{lee2020drit++}, although we deployed the same network architecture. BicycleGAN~\cite{zhu2017toward} achieved larger diversity on all tasks by adding noise to all decoders through the U\_Net~\cite{ronneberger2015u}, but the tradeoffs are worse visual results, due to the larger noise being directly added to the last generative layer.
 In contrast, following in MUNIT~\cite{huang2018multimodal}, we only added noise to the middle layers of generation, through AdaIN.
 
 \begin{figure*}[tb!]
	\centering
	\setlength{\abovecaptionskip}{0.cm}
	\setlength{\belowcaptionskip}{-0.cm}
	\includegraphics[width=\linewidth]{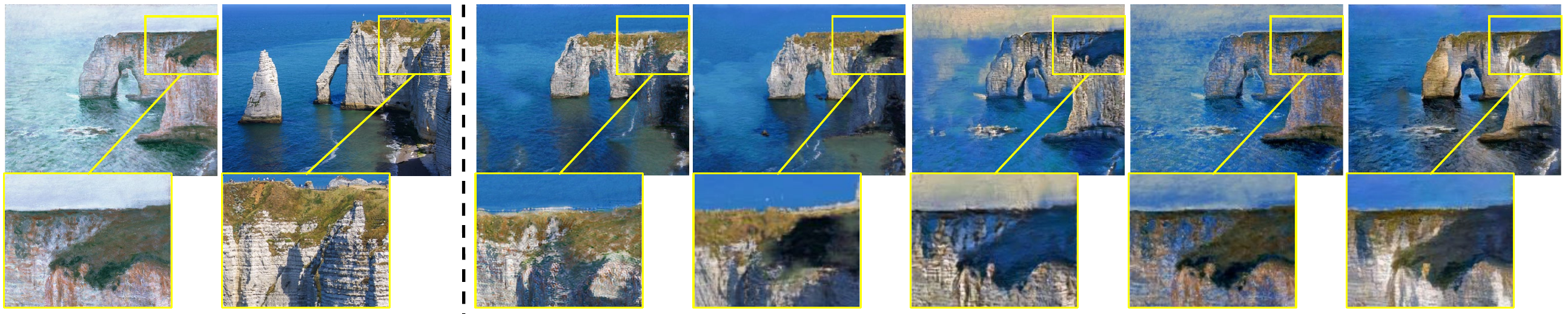}
	\begin{picture}(0,0)
		\put(-230,5){\footnotesize Input}
		\put(-150,5){\footnotesize Target}
		\put(-80,5){\footnotesize FSeSim}
		\put(-10,5){\footnotesize CUT~\cite{park2020cut}}
		\put(50,5){\footnotesize Gatys \etal~\cite{gatys2016image}}
		\put(125,5){\footnotesize WCT$^2$~\cite{yoo2019photorealistic}}
		\put(190,5){\footnotesize STRORSS~\cite{kolkin2019style}}
	\end{picture}
	
	\caption{\textbf{High-resolution painting to photorealisitc image} on single-image translation. Zoom in to see the details.}
	\label{fig:results_sin_img_real}	
\end{figure*}

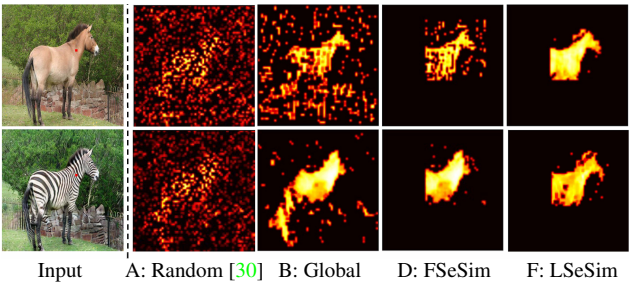
\begin{figure}[tb!]
	\centering
	\setlength{\abovecaptionskip}{0.cm}
	\setlength{\belowcaptionskip}{-0.cm}
	\includegraphics[width=\linewidth]{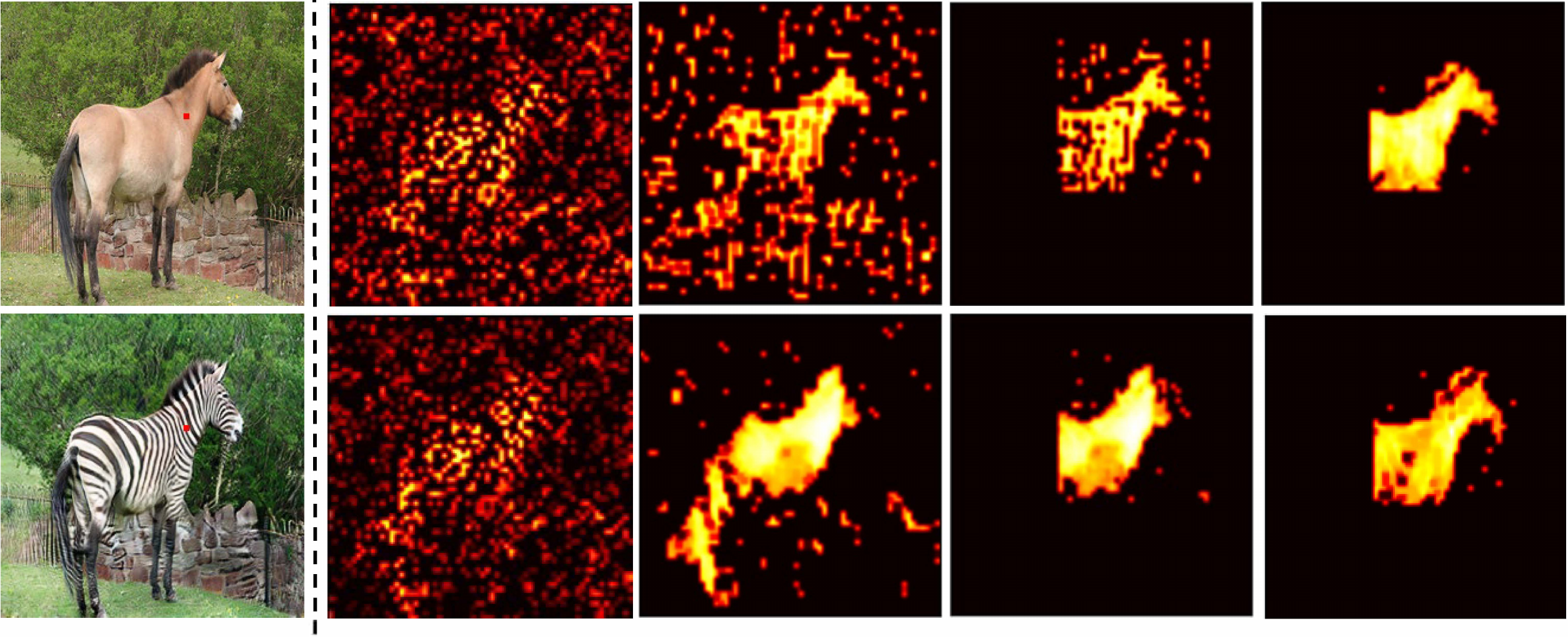}
	\begin{picture}(0,0)
		\put(-105,5){\footnotesize Input}
		\put(-72,5){\footnotesize A: Random~\cite{kolkin2019style}}
		\put(-14,5){\footnotesize B: Global}
		\put(31,5){\footnotesize D: FSeSim}
		\put(80,5){\footnotesize F: LSeSim}
	\end{picture}
	\caption{\textbf{Ablation study on self-similarity maps}. A, B, D and F correspond to the settings in Table~\ref{table:ablation}, respectively.} 
	\label{fig:results_ablation}	
\end{figure}

In Fig.~\ref{fig:results_mul_mod}, we show qualitative comparisons of our method to MUNIT~\cite{huang2018multimodal} on \emph{winter $\rightarrow$ summer}, and \emph{night $\rightarrow$ day} tasks. As can be seen, our model not only generated higher quality translated results, but also produced more diverse solutions for these multi-modal tasks. We believe this is because the formulation of our F/LSeSim will only maintain structural fidelity, and does not impose penalties on appropriate appearance modifications in the target domain.

\paragraph{Ablation Experiments.} To understand the influence of different components for the proposed spatially-correlative loss, we ran a number of ablations. The quantitative results are reported in Table~\ref{table:ablation} for both single- and multi-modal image translation. In this table, \textbf{row A} shows the performance of the baseline method~\cite{kolkin2019style} which utilizes self-similarity as content loss. However, it calculates the similarity using random sampled features and does not have an explicit connection to spatial structure. \textbf{Row B} is a global attention map. While this version performed well and ran faster by avoiding sampling, it has two main limitations. First, the original global attention module is memory intensive and \emph{cannot} be applied to multiple scales nor to large feature spaces. Second, as evident from Fig.~\ref{fig:results_ablation}, the spatially distant correlation is essentially noise (as is also the case for the Random baseline of \cite{kolkin2019style}), which is detrimental to the results. Compared to the global version, \textbf{row C} largely improved the performance as clearer shapes are captured in the local patches. In \textbf{row D}, we applied local attention to multiple layers with a fixed path size. This results in the spatially-correlative maps having different receptive fields, which further improves the performance. \textbf{Row E} replaces the $l_1$ distance with cosine distance. While the improvement in image quality is not obvious, the diversity scores increased substantially. This is due to the cosine similarity supporting only the correlation between the two spatially-correlative maps without encouraging the maps to be fully same. \textbf{Row E} shows the performance of the full model (same as in Tables~\ref{table:single_mod} and~\ref{table:multi_modal}), where LSeSim of \textbf{row F} improved on many metrics, and had better visual results.

\subsection{\emph{Single-Image} Unpaired Image Translation}

To further test the generalization ability, we applied the FSeSim to a high-resolution single-image translation task. Here, only one source and one target image are provided for training, but they are unpaired. This task is conceptually similar to the style transfer~\cite{gatys2016image,johnson2016perceptual,luan2017deep}, except that here we trained a SinGAN-like~\cite{shaham2019singan,InGAN} model that captures the distribution of a single image through the adversarial learning, rather than using a fixed style loss \cite{johnson2016perceptual}.

\paragraph{Implementation details.} The single-image translation setting is based on the latest CUT~\cite{park2020cut} method, except that the PatchNCE loss is replaced by our FSeSim loss. In detail, the StyleGAN2-based generator and discriminator~\cite{karras2020analyzing} with the gradient penalty~\cite{mescheder2018training,karras2019style} are used. To further increase simplicity, we removed the identity loss in CUT~\cite{park2020cut}, and only used a GAN loss in conjunction with the proposed FSeSim loss to assess the appearance and structure separately. As $64\times64$ crops have to be taken from a high-resolution image for training here, it becomes less useful to further subsample ``positive'' and ``negative'' patches. Therefore, we only use our FSeSim to train the model, without using LSeSim with contrastive loss.

\paragraph{Results.} In Fig.~\ref{fig:results_sin_img_real}, we show qualitative results from the CUT~\cite{park2020cut} paper on the \emph{painting}$\rightarrow$\emph{photo} task. As evident in the highlighted regions, our model generated not only higher quality results, but they were also closer to the target image style than existing methods, including classical style transfer models, such as WCT$^2$~\cite{yoo2019photorealistic} and STRORSS~\cite{kolkin2019style}, as well as the latest single-image translation CUT~\cite{park2020cut} model.  

\section{Conclusion}

We introduced F/LSeSim, a new structure consistency loss that focuses only on spatially-correlative relationships, without regard to visual appearances. Our F/LSeSim is naturally suitable for tasks that require structure consistency, and can be easily applied to existing architectures. We demonstrated its generality to various unpaired I2I translation tasks, where a simple replacement of the existing content losses with F/LSeSim led to solid performance improvements. 

\paragraph{\bf Acknowledgements} This research was supported by the National Research Foundation, Singapore under its International Research Centres in Singapore Funding Initiative. This research was also supported by the Monash FIT Start-up Grant. Any opinions, findings and conclusions or recommendations expressed in this material are those of the author(s) and do not reflect the views of National Research Foundation, Singapore. 

{\small
\bibliographystyle{ieee_fullname}
\bibliography{egbib}
}

\appendix\onecolumn
\renewcommand{\theequation}{\thesection.\arabic{equation}}
\setcounter{equation}{0}
\renewcommand{\thefigure}{\thesection.\arabic{figure}}
\setcounter{figure}{0}
\renewcommand{\thetable}{\thesection.\arabic{table}}
\setcounter{table}{0}
\newpage


\section{Additional Examples}\label{results}

\subsection{Additional comparisons}

\begin{figure*}[htb!]
	\centering
	\setlength{\abovecaptionskip}{0.cm}
	\setlength{\belowcaptionskip}{-0.cm}
	\includegraphics[width=\linewidth]{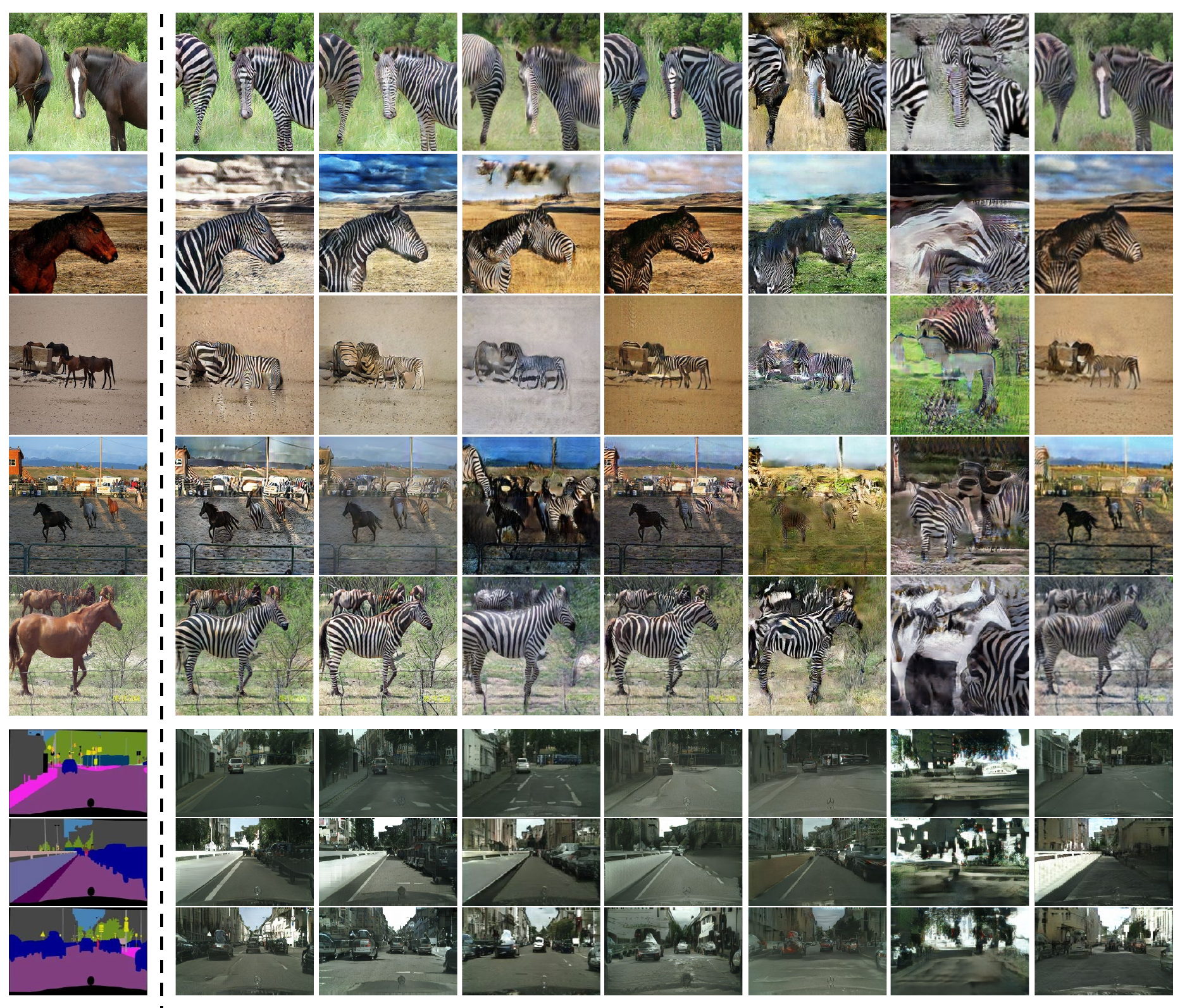}
	\begin{picture}(0,0)
		\put(-225,6){\footnotesize Input}
		\put(-160,6){\footnotesize LSeSim}
		\put(-100,6){\footnotesize FSeSim}
		\put(-37,6){\footnotesize CUT\cite{park2020cut}}
		\put(12,6){\footnotesize CycleGAN\cite{zhu2017unpaired}}
		\put(78,6){\footnotesize MUNIT\cite{huang2018multimodal}}
		\put(128,6){\footnotesize DistanceGAN\cite{benaim2017one}}
		\put(200,6){\footnotesize GcGAN\cite{fu2019geometry}}
	\end{picture}
	
	\caption{\textbf{More qualitative results on single-modal image-to-image translation.} Here, we show the same examples as in CUT paper~\cite{park2020cut}. Our model produces better visual results in most cases, changing the appearance while preserving the structure well. Please zoom in to see fine image detail. } 
	\label{fig:appendix_sin_mod}	
\end{figure*}
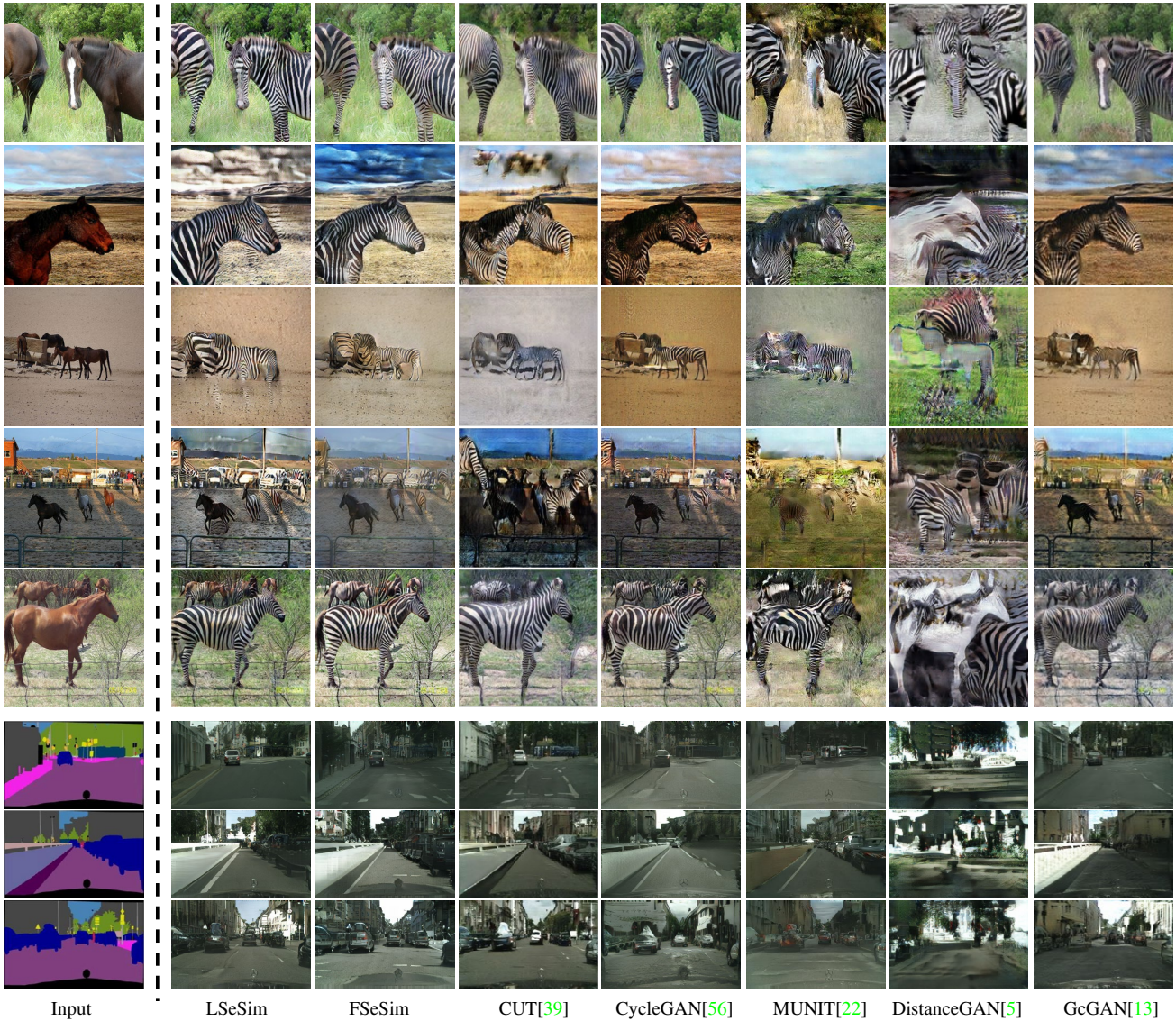

In Fig.~\ref{fig:appendix_sin_mod}, we show additional comparison results for \emph{horse $\rightarrow$ zebra} and \emph{label $\rightarrow$ image}. This is an extension of Fig.~\ref{fig:results_sin_mod} in the main paper. Here, all examples shown are those chosen in the CUT paper~\cite{park2020cut}. When compared to  CUT~\cite{park2020cut} and CycleGAN~\cite{zhu2017unpaired}, although we used the same setting for training, our model produces better visual results, even under challenging conditions, \eg the results in rows 3 and 4. 

\begin{figure*}[tb!]
	\centering
	\setlength{\abovecaptionskip}{0.cm}
	\setlength{\belowcaptionskip}{-0.cm}
	\includegraphics[width=\linewidth]{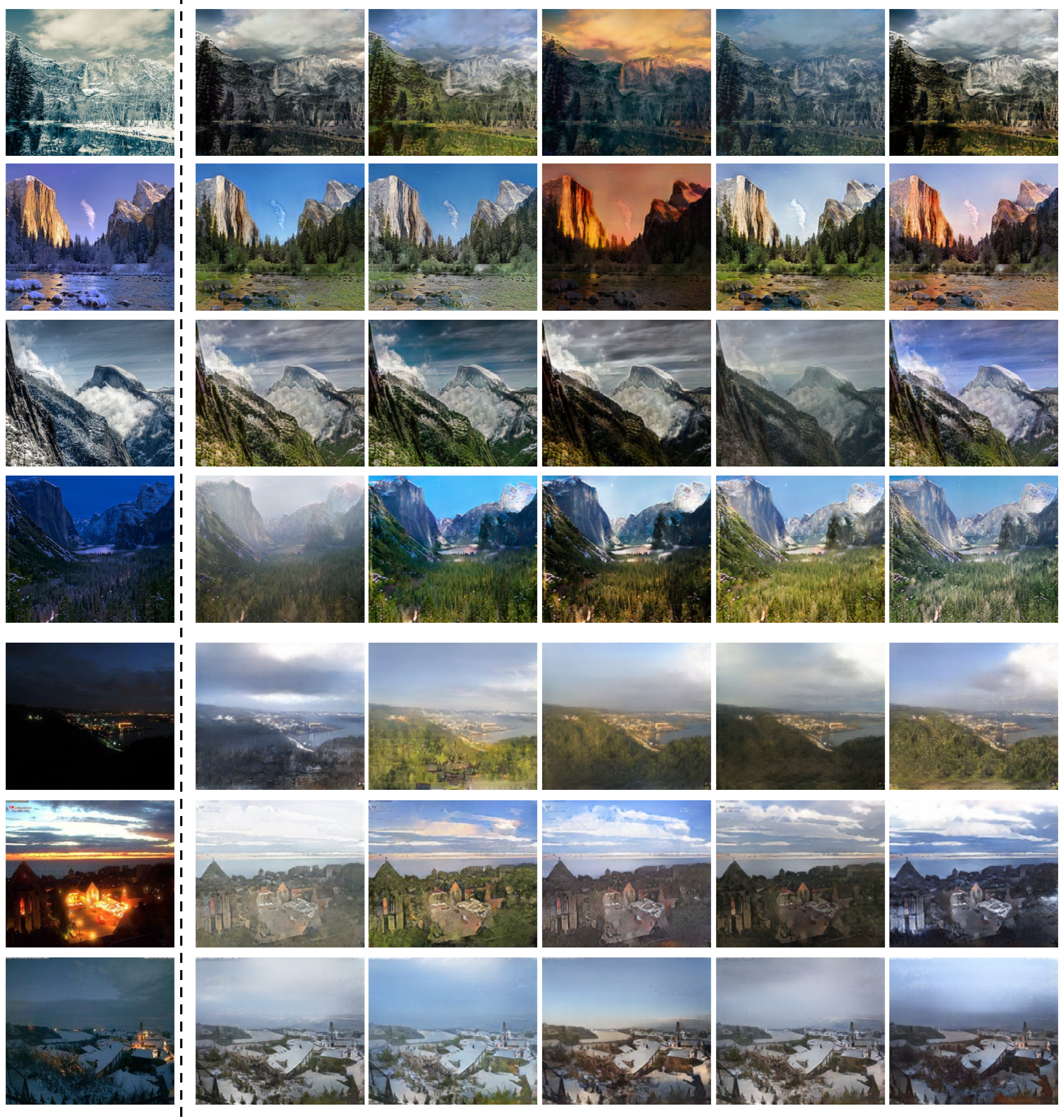}
	\begin{picture}(0,0)
		\put(-215,6){Input}
		\put(32,6){LSeSim}
	\end{picture}
	
	\caption{\textbf{More qualitative results on multi-modal image-to-image translation.} To demonstrate our model is suitable for the task that requires multiple and diverse results, here we show various sampled results on \emph{winter $\rightarrow$ summer} and \emph{night $\rightarrow$ day} task. This is an extension of Fig.~\ref{fig:results_mul_mod} in the main paper. As we can see, for \emph{winter $\rightarrow$ summer}, our model produces various color and lighting, but all keeping within the range of appearances for the summer domain. The obvious appearance modification is also presented in the \emph{night $\rightarrow$ day} results. } 
	\label{fig:appendix_mul_mod}	
\end{figure*}

In Fig.~\ref{fig:appendix_mul_mod}, we present multiple  diverse results for the multi-modal image-to-image translation task. This is an extension of Fig.~\ref{fig:results_mul_mod} in the main paper. 

\newpage
\section{Experimental Details}\label{training}

\subsection{Architecture details}

To demonstrate the generality of our proposed F/LSeSim losses, we intentionally applied them to multiple existing architectures for various image-to-image translation tasks, including CycleGAN~\cite{zhu2017unpaired} for \emph{single-modal} image translation, MUNIT~\cite{huang2018multimodal} for \emph{multi-modal} image translation, and CUT~\cite{park2020cut} for \emph{single-image} translation. The architecture details can be found in \href{https://github.com/junyanz/pytorch-CycleGAN-and-pix2pix}{https://github.com/junyanz/pytorch-CycleGAN-and-pix2pix}, \href{https://github.com/NVlabs/MUNIT}{https://github.com/NVlabs/MUNIT}, and \href{https://github.com/taesungp/contrastive-unpaired-translation}{https://github.com/taesungp/contrastive-unpaired-translation}, respectively. We use their generator and discriminator architectures to evaluate the proposed F/LSeSim losses.

Our structure representation network is based on the VGG16~\cite{simonyan2014very}, without the batch normalization layer. Our global version \textbf{Row B} of Table~\ref{table:ablation} in the main paper is computed using the features of layer ``relu3\_1'' from a fixed ImageNet pre-trained model. The feature size is $64\times64$.  The local version \textbf{Row D} and \textbf{Row E} of Table~\ref{table:ablation} is constructed using the features of layer ``\texttt{relu}3\_1'' and ``\texttt{relu}4\_1'', with parch size $32\times32$. The features used in LSeSim also come from the layer ``\texttt{relu}3\_1'' and ``\texttt{relu}4\_1''. However, we further added two convolution layers, implemented as $1\times1$ kernels, to select these features for a specific task. 

\subsection{Evaluation details}

\paragraph{Fr\'echet Inception Distance (FID)~\cite{heusel2017gans}} is computed by measuring the mean and variance distance of the generated and real images in a deep feature space. Here, we used the default setting of \href{https://github.com/mseitzer/pytorch-fid}{https://github.com/mseitzer/pytorch-fid} to compute the FID score. For the single-modal image translation, we directly compared the mean and variance of generated and real sets. As for the multi-modal image translation, we sampled 19 times for each test image. Therefore, we computed the FID for each sampled set and averaged the scores to get the final result. 

\paragraph{Semantic segmentation accuracy (on the Cityscapes dataset)} is computed following SPADE~\cite{park2019semantic} and CUT~\cite{park2020cut}. We first trained the segmentation network DRN using the publicly-available code at \href{https://github.com/fyu/drn}{https://github.com/fyu/drn} using the default setting, except that the input resolution is $256\times256$ to be consistent to our translated output. We then conducted semantic segmentation on the generated images and evaluated the results by comparing the labels to the ground-truth paired semantic maps.

\paragraph{Average LPIPS distance~\cite{zhang2018unreasonable}} measures the distance between two images in a feature domain, which correlates well to human perceptual recognition. In our paper, we used the code at \href{https://github.com/richzhang/PerceptualSimilarity}{https://github.com/richzhang/PerceptualSimilarity} to evaluate the perceptual similarity of two images. If the distance is large, the appearance diversity is high. Following BicycleGAN~\cite{zhu2017toward} and MUNIT~\cite{huang2018multimodal}, we sampled 100 images with 19 pairs to obtain multiple and diverse results. We then compared the distance for each paired generated results and averaged all distances to get the average LPIPS score. 

\paragraph{Density and Coverage (D\&C)~\cite{ferjad2020icml}} is the latest metric for simultaneously judging the diversity and fidelity of generated results. Given the generated and real images, we also encoded them to a deep feature domain and then utilized the manifold method to compute the precision and recall using the features of generated images and the real images. Similar to LPIPS, we first sampled 19 pairs for each image, and then used the code at \href{https://github.com/mseitzer/pytorch-fid}{https://github.com/mseitzer/pytorch-fid} to extract the features of real and generated images. Finally, we fed these 1900 generated samples and real features with 2048-dimensions to the PRDC function provided in \href{https://github.com/clovaai/generative-evaluation-prdc}{https://github.com/clovaai/generative-evaluation-prdc} to calculate these scores.

\end{document}